\pdfoutput=1
\documentclass[sigconf,nonacm]{acmart}
\usepackage{pifont}
\usepackage[ruled,vlined,linesnumbered]{algorithm2e}
\usepackage{multirow}
\usepackage{tikz}
\usepackage{makecell}
\usepackage{float}
\usepackage{subcaption}

\renewcommand\footnotetextcopyrightpermission[1]{}
\AtBeginDocument{%
  }

\begin{document}

\title{Cross-Layer Misalignment Detection in Agent Skills: A Progressive Loading-Aware Contrastive Learning Approach}

\newcommand{\iuaffiliation}{%
  \affiliation{%
    \institution{Indiana University Bloomington}
    \city{Bloomington}
    \state{Indiana}
    \country{USA}
  }%
}

\author{Chengjun Zhang}
\email{cz1@iu.edu}
\iuaffiliation

\author{Yang Gao}
\email{gaoyang@iu.edu}
\iuaffiliation

\author{Jianna Hur}
\email{jhur06@iu.edu}
\iuaffiliation

\author{Jingjing Zhang}
\email{jjzhang@iu.edu}
\iuaffiliation

\author{Sagar Samtani}
\email{ssamtani@iu.edu}
\iuaffiliation

\begin{abstract}
Large language model (LLM) agents are increasingly extended through Agent Skills, reusable artifacts that package natural-language metadata, procedural instructions, and execution-time resources for runtime use. As open-source skill marketplaces expand, users and agents increasingly rely on brief metadata to select third-party skills, making it difficult to detect inconsistencies between a skill’s description and its true behavior, a problem we call \emph{cross-layer misalignment}. To address this issue, we propose \textbf{Progressive Loading-Aware Hierarchical Contrastive Learning (PL-HCL)}, an LLM-based framework that detects misalignment by modeling the layered structure of Agent Skills and learning cross-layer consistency. Using a normalized corpus of over 264{,}000 open-source skills and a human-verified challenge set, PL-HCL improves Macro-F1 from approximately 0.45 for unadapted baselines to 0.87--0.89 across evaluated LLM backbones. This approach offers an effective screening tool for users and operators, as well as design principles for detecting inconsistencies in layered digital artifacts.
\end{abstract}




\maketitle


\section{Introduction}

Large Language Models (LLMs) are evolving from single-turn text generators into agentic Artificial Intelligence (AI) systems capable of reasoning, tool use, and interaction with external environments~\cite{yao2023react,schick2023toolformer}. Agent Skills have become essential supply-side components of agentic AI systems to extend an agent's capabilities ~\cite{anthropic2025agentskills,li2026skillsbench}. \emph{Agent Skills} encapsulate reusable procedural knowledge into loadable modules, which typically include natural language instructions, scripts, and resources to support specific tasks or workflows~\cite{anthropic2025agentskills}.

In contrast to standard prompts or standalone tools, Agent Skills are hierarchical artifacts in which a surface metadata layer sits above layers of instructions, code, files, and executables. Users and agents access metadata before progressively accessing deeper components. This progressive loading mechanism introduces the risk that the metadata used for skill selection may not correspond to the behaviors actually supported or enabled by the underlying components ~\cite{greshake2023notwhat,zhan2024injectagent,debenedetti2024agentdojo}. For example, a skill may claim in its metadata to be a harmless productivity assistant while containing prompt injection, credential leakage, data exfiltration, or insecure command execution within its instruction or script layers~\cite{liu2026agentskillswild,liu2026maliciousagentskills}. Conversely, a skill may claim to support a specific capability in its description, yet fail to implement it within its instructions or resources. This reflects a broader documentation-implementation consistency problem, where natural-language descriptions can diverge from the executable or procedural artifacts they are intended to summarize~\cite{tan2012icomment,zhou2017aristotle,liu2018neural}. In Agent Skill packages, such inconsistency can be described as \emph{cross-layer misalignment.} In this case, user-facing claims may diverge from the behaviors supported, requested, or enabled by the underlying instructions, resources, and executable components.

This mismatch warrants attention because trust decisions regarding Agent Skills are frequently made before the package is fully inspected or executed. Marketplace descriptions, repository metadata, and skill names influence the selection process. However, the underlying instruction and resource layers that dictate behavior may not be accessible until later loading or execution stages. Consequently, the trustworthiness of Agent Skills cannot be assessed solely by examining the security of a single layer or by evaluating task success rates after execution. Past research has shown that agent and skill evaluation often focuses either on security vulnerabilities in specific artifacts or on downstream task performance, while cross-layer consistency between surface claims and deeper package evidence remains a distinct evaluation target~\cite{debenedetti2024agentdojo,liu2026maliciousagentskills,li2026skillsbench,mialon2023gaia,han2026sweskillsbench}. Instead, it is necessary to determine whether a consistent behavioral contract exists across the different layers prior to execution.

Current evaluations of agentic AI focus primarily on runtime performance, including task success rates, reasoning-action trajectories, web browsing, multimodal reasoning, or tool proficiency~\cite{mialon2023gaia,yao2023react}. Although these paradigms are essential for understanding the execution process, they generally assume that the relevant tools or skills have already been selected and loaded~\cite{mialon2023gaia,li2026skillsbench,han2026sweskillsbench,debenedetti2024agentdojo}. The key question is whether the components within a skill package are internally consistent prior to execution. This ``pre-execution" setting requires the model to compare heterogeneous forms of evidence, specifically metadata, instructions, and resources. Although general LLMs or cybersecurity LLMs may detect suspicious text or code patterns, they have not been explicitly trained to verify whether high-level claims are genuinely supported by the evidence contained in the underlying package components. In light of these issues, we propose a Progressive Loading-Aware Hierarchical Contrastive Learning (PL-HCL) framework that aims to evaluate cross-layer misalignment in Agent Skills before execution. PL-HCL treats each skill package as a layered artifact consisting of metadata, instructions, and resources, and trains the model to learn whether user-facing claims are supported by deeper package evidence. The framework integrates two key components:

\begin{enumerate}
    \item A two-stage continued pretraining process that adapts base LLMs to the vocabulary, formatting, and progressive structure of Agent Skill packages. The first stage trains on short metadata-instruction views, and the second stage trains on full package views that include resources.
    \item A progressive loading-aware hierarchical contrastive learning process learns cross-layer consistency by contrasting aligned same-skill layers with metadata-swapped and corrupted variants. This design directly targets the mismatch between surface claims and deeper behavioral evidence.

\end{enumerate}
This paper makes several key contributions. First, we construct a large-scale open-source Agent Skill corpus and a human-verified challenge set for aligned/misaligned skill evaluation, enabling systematic study of skill-level trustworthiness before execution. Second, we introduce PL-HCL, a contrastive pretraining framework that learns cross-layer consistency from aligned packages and synthetic pairs. Finally, we evaluate base LLMs, two-stage continued pretraining, and PL-HCL across general and cybersecurity backbones to demonstrate that skill-format adaptation alone is insufficient while explicit cross-layer contrastive learning substantially improves misaligned-class detection.

The remainder of this paper is organized as follows. We first review prior work on Agent Skill security, continued pretraining and curriculum adaptation, and contrastive learning for hierarchical and multi-view representations. Second, we identify research gaps and present the research questions for this study. Third, we define cross-layer misalignment and describe the dataset. We then present the PL-HCL framework and report the experimental results. Finally, we discuss implications for trustworthy Agent Skill evaluation and conclude with limitations and future directions.

\section{Related Work}
\label{sec:related-work}
This section reviews three streams of related research that motivate and ground our problem formulation and proposed method: \textbf{1) work on tool-augmented agents and Agent Skill security} explains why reusable skill components create a new trustworthiness surface for agentic AI; \textbf{2) continued pretraining and curriculum adaptation} motivate our use of staged skill-format adaptation before contrastive learning, and \textbf{3) contrastive and hierarchical representation learning} provide the methodological foundation for learning cross-layer consistency within agent skills.

\subsection{Tool-Augmented Agents and Agent Skill Security}
\label{sec:rw-skill-trust}
\subsubsection{Tool-Augmented Agent Evaluation}
Research on tool-augmented agents explores how LLMs extend their reasoning and task-completion capabilities through external tools, APIs, and environment actions. For example, ReAct demonstrated an agent pattern that interleaves reasoning traces with actions, enabling LLMs to invoke external information or tools during task execution~\cite{yao2023react}. Toolformer further investigated how LLMs can learn to call APIs at appropriate junctures, thereby integrating tool use into model behavior~\cite{schick2023toolformer}. As agent systems shift from isolated text generation to external interaction, evaluation has increasingly focused on agent performance in complex task environments. GAIA evaluates the task-completion capabilities of general AI assistants in dimensions such as reasoning, web browsing, multimodal understanding, and tool-use proficiency~\cite{mialon2023gaia}. Recent red-teaming work further shows that autonomous agents equipped with persistent memory, communication channels, file-system access, or shell execution can produce failures that involve security, privacy, and governance risks. These risks include information disclosure, destructive actions, identity spoofing, and partial system takeover~\cite{shapira2026agentschaos}. Such research demonstrates that evaluating agentic AI requires attention not only to the quality of the final answer but also to the broader trustworthiness risks arising from the interaction between agents and external capabilities. Therefore, we review the characteristics of Agent Skill.

\subsubsection{Agent Skills as Reusable Evaluation Artifacts}
As indicated in the introduction, Agent Skills encapsulate procedural knowledge into reusable packages, enabling agents to load task-specific capabilities at inference time~\cite{anthropic2025agentskills,li2026skillsbench}. Unlike a single API endpoint or function call, Agent Skill packages typically comprise marketplace metadata, natural-language instructions, scripts, configuration files, dependencies, and resources~\cite{anthropic2025agentskills}. Thus, Agent Skills are not merely external interfaces invoked by agents. Instead, skills are structured artifacts containing claims, instructions, and executable or resource-based evidence. Recent work has begun to investigate Agent Skills as independent objects of evaluation. For example, SkillsBench examines whether curated or self-generated skills enhance agent performance across tasks, finding that curated skills increase average pass rates, but their effectiveness depends on the domain and task~\cite{li2026skillsbench}. Such benchmarks isolate agent skills from general tool-use settings and emphasize the impact of the skill package itself on agent performance. For skill-based agent ecosystems, this demonstrates that reusable skill packages are both auxiliary materials and evaluation artifacts capable of substantially altering agent behavior and task outcomes.

\subsubsection{Security and Trustworthiness Risks in Agent Skills}
As Agent Skills are increasingly used to extend agent capabilities, security studies have begun to systematically analyze their risks. Liu et al.~\cite{liu2026maliciousagentskills} conducted a security analysis of a large-scale set of skills from community registries, finding that malicious skills can facilitate credential theft, remote code execution, agent manipulation, and other attack behaviors through natural-language instructions and helper scripts. Other studies have also highlighted issues such as prompt injection, credential leakage, data exfiltration, privilege escalation, unsafe code execution, and hidden malicious behavior in open-source skills~\cite{liu2026agentskillswild,schmotz2026skillinject,jin2026skillsafetybench}. Taken together, these studies show that Agent Skills play a vital role in establishing trustworthiness in agentic AI from the supply side. Evaluating the structure, content, and potential security issues of reusable skill packages is a crucial prerequisite for understanding and improving the trustworthiness of agentic AI within skill-based agent ecosystems. However, the evolving and hierarchical nature of Agent Skills necessitates appropriate training and representation learning strategies. 

\subsection{Continued Pretraining and Curriculum Adaptation}
\label{sec:rw-cpt}
Agent Skill packages are layered artifacts comprising metadata, instructions, and resources. Metadata and skill descriptions are often surface-level capability claims, while the underlying \texttt{SKILL.md} file, resources, and executables are longer, heterogeneous, and contain clearer behavioral evidence. Consequently, models need to adapt to a skill's high-level descriptions before engaging with the more complex, resource-level information. Curriculum learning has emerged as a training strategy in which a model learns from easy to hard samples, and holds significant promise for modelling Agent Skill structures. Early work indicated that models could first learn fundamental structures from simpler or clearer examples before gradually transitioning to more complex ones, thereby improving optimization and generalization~\cite{bengio2009curriculum}. Subsequent research extended into self-paced learning and competence-based curriculum learning. These approaches allow models to encounter data with varying difficulty levels, determined by sample difficulty or model competence, throughout training stages~\cite{kumar2010selfpaced,platanios2019competence}. This body of work demonstrates that the training process is not merely a matter of repeated sampling from a fixed data distribution; the training order itself serves as a crucial learning signal, helping the model progressively adapt to complex structures.

While continued pretraining addresses different issues than curriculum learning, the two are complementary in the context of model adaptation. Continued pretraining typically involves further training a pre-existing language model on domain-specific or task-relevant corpora, enabling it to adapt to the target domain's vocabulary, style, and discourse patterns prior to downstream supervised learning. Gururangan et al.~\cite{gururangan2020dont} conducted a systematic study of domain-adaptive pretraining (DAPT) and task-adaptive pretraining (TAPT), demonstrating that adaptive pretraining can enhance downstream performance across various domains, such as biomedicine, computer science, news, and reviews. Related works have also applied domain-adaptive methods to specialized text-analysis settings, including legal, scientific, biomedical, and news-analysis tasks \cite{lee2020biobert,beltagy2019scibert,chalkidis2020legalbert,zhang2026domainadaptive}. 
These studies collectively indicate that model adaptation often requires considering both \emph{what data} the model sees and \emph{in what order} it sees them. In the context of Agent Skill analysis, this requires a strategy to carefully represent the hierarchical and multi-faceted structure of skills. 

\subsection{Contrastive Learning for Hierarchical and Multi-View Representations}
\label{sec:rw-contrastive}

For Agent Skill packages, metadata, instructions, and resources can be seen as distinct layers or views of the same skill artifact. Metadata provides user-facing capability claims, instructions offer agent-facing procedural guidance, and resources, along with executable artifacts, serve as evidence that is closer to actual behavior. Therefore, hierarchical and multi-view contrastive learning provide a methodological foundation for cross-layer representation learning. Contrastive learning is a class of self-supervised representation learning that seeks to pull related views or semantically similar samples closer together while pushing unrelated samples further apart. A prevailing approach is SimCLR, which constructs positive pairs from different augmentations of the same image and employs a contrastive objective to learn transferable visual representations~\cite{chen2020simclr}. Supervised contrastive learning further incorporates label information into the contrastive objective, bringing samples of the same class closer together in the representation space while increasing the separation between samples of different classes~\cite{khosla2020supcon}. In the realm of language-image representation learning, CLIP demonstrates the effectiveness of contrastive objectives for cross-modal alignment and transferable representation learning by aligning image-text pairs~\cite{radford2021clip}. These works show that contrastive learning, when using carefully defined positive and negative samples, can shift representation learning away from a singular focus on reconstruction or classification and toward modeling relationships.

Beyond general contrastive learning, multi-view contrastive learning focuses on how different views of the same underlying object share semantic factors. Contrastive Multiview Coding (CMC) treats different sensory channels as complementary sources of information, learning view-invariant representations by maximizing cross-view agreement for the same instance~\cite{tian2020cmc}. Similarly, CLIP views images and text as distinct modalities of the same semantic content, learning aligned multimodal embeddings through large-scale contrastive pretraining~\cite{radford2021clip}. These studies show that contrastive learning facilitates learning both individual representations and shared semantics when objects have multiple views.

Hierarchical contrastive learning extends the contrastive objective to data characterized by hierarchies or multi-level structures. In hierarchical text classification, label taxonomies provide a class-level hierarchy; models can leverage parent-child label relationships, label paths, or hierarchy-aware positive samples to learn text representations that better align with the underlying label structure. HGCLR integrates the label hierarchy into the text encoder, learning hierarchy-aware representations through hierarchy-guided positives~\cite{wang2022hgclr}. Hierarchical multi-label contrastive learning further leverages multi-level labels and label relations to preserve the class hierarchy, ensuring the representation space reflects the structural relationships between different labels~\cite{zhang2022usealllabels}. In relation extraction and recommendation settings, hierarchical contrastive objectives are also employed to combine global structure with local interactions, thereby learning multi-granularity representations~\cite{li2022hiclre,wu2023hicon}. These studies show that contrastive learning supports structure-aware representations, with positive and negative pairs defined by labels, modalities, views, or hierarchical relations, not just data augmentation. Within the context of Agent Skill analysis, these approaches can help models learn shared skill semantics by aligning layers belonging to the same skill, while also learning to detect signals of inconsistency between layers through mismatched or corrupted layer combinations.

\section{Research Gaps and Questions} 
\label{sec:research-gaps}  
We identified several research gaps from our literature review. First, although research on tool-augmented agents and Agent Skill security is expanding, current evaluation frameworks offer limited mechanisms for assessing the trustworthiness of a skill package prior to execution. Most agentic AI benchmarks focus on evaluating task completion, reasoning processes, or tool-use proficiency only after tools or skills have been selected and loaded. Consequently, there remains an unresolved challenge in evaluating the skill artifact itself at the stage when users or agents must make trust decisions based solely on partial or surface-level information. Second, existing Agent Skill security research has identified specific risks, including prompt injection, credential leakage, data exfiltration, unsafe code execution, and concealed malicious behavior. However, these studies predominantly address vulnerabilities or malicious intent, whereas Agent Skill trustworthiness also relies on the alignment between user-facing descriptions and the underlying instruction and resource layers. As a result, a skill may be misaligned without being overtly malicious, such as when its metadata exaggerates or inaccurately represents capabilities not supported by the underlying package. Third, although continued pretraining helps LLMs learn skill package formats, it does not train models to compare claims across different layers. Finally, contrastive and hierarchical methods leverage structure from labels or graphs, but Agent Skills require a unique structural signal. Their hierarchy arises from progressive loading, with metadata first and instructions or resources providing deeper behavioral evidence. Based on these gaps, we pose the following research questions: 
\begin{itemize}     
\item \textbf{RQ1:} How can cross-layer misalignment in Agent Skills be formalized as a pre-execution, artifact-level evaluation problem?      
\item \textbf{RQ2:} To what extent can base general and cybersecurity LLMs detect cross-layer misalignment without skill-specific adaptation?      
\item \textbf{RQ3:} How does two-stage continued pretraining improve misalignment detection?      
\item \textbf{RQ4:} How does progressive loading-aware hierarchical contrastive learning improve misaligned-skill detection beyond base LLMs and CPT-only checkpoints? \end{itemize}




\section{Progressive Loading-Aware Hierarchical Contrastive Learning (PL-HCL) Framework for Detecting Misaligned Agent Skills}
\label{sec:method}

\begin{figure*}[t]
    \centering
    \includegraphics[width=\linewidth]{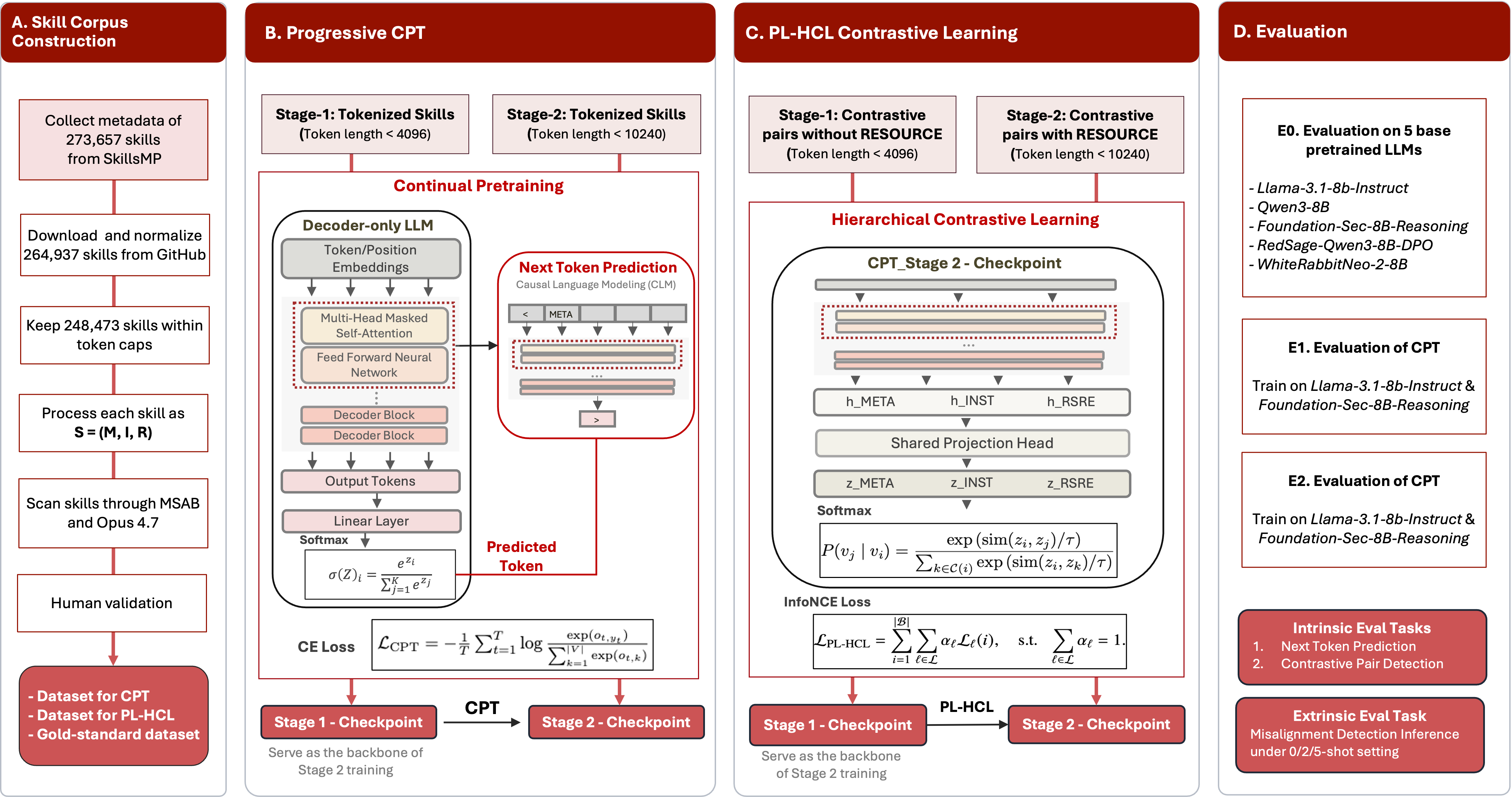}
    \caption{Overview of the PL-HCL framework. Normalized skill packages are represented as metadata \(M\), instructions \(I\), and resources/executables \(R\). Two-stage CPT adapts the base LLM to short and full skill views. PL-HCL then learns cross-layer consistency by contrasting aligned packages with swapped and corrupted variants.}
    \label{fig:plhcl-framework}
\end{figure*}

We propose a PL-HCL framework for Agent Skill misalignment detection, illustrated in Figure~\ref{fig:plhcl-framework}. The proposed framework consists of four major components: (1) public Agent Skill data is collected and processed; (2) two-stage continued pretraining adapts the base LLM to these packages; (3) progressive loading-aware hierarchical contrastive learning then distinguishes between aligned and misaligned skill layers; and (4) the model determines whether user-facing metadata is supported by underlying instructions and resources. Each component is described in the following subsections.

\subsection{Skill Corpus Construction}
\label{sec:data-acquisition}

We build an open-source Agent Skill corpus by linking public records from SkillsMP~\cite{skillsmp2026}, the largest open-source Agent Skills platform, to their GitHub repositories and normalizing each retrieved package into a layered representation \(S=(M,I,R)\): the metadata layer \(M\) contains user-facing fields such as the skill name, description, categories, tags, repository identifiers, and declared resource references; the instruction layer \(I\) contains agent-facing procedural guidance; and the resource layer \(R\) contains scripts, configuration files, dependencies, examples, assets, and other supporting files. From 273,657 SkillsMP catalog entries, 264,937 packages were successfully downloaded and normalized; after applying the 4,096-token metadata--instruction budget and the 10,240-token full-package budget, 248,473 packages fit at least one training stage and form the unlabeled training corpus, while the held-out Challenge Set is excluded from both CPT and PL-HCL training.\footnote{Detailed corpus construction, token-budget coverage, and filtering statistics are provided in the supplementary materials.}

This representation preserves the distinction between user-facing claims and deeper package evidence, supporting our pre-execution, artifact-level task: the detector observes package artifacts available at inspection time and assesses whether user-facing metadata is supported by the instruction and resource layers. We do not verify runtime-only behaviors unless the package provides observable evidence.

\subsection{Progressive CPT}

\label{sec:cpt}

The prediction target of our proposed approach is whether the layers of a skill package are mutually consistent: 0 if aligned, 1 if misaligned. A skill is considered misaligned when its metadata describes a capability or intent not supported by the instruction or resource layers. We consider two forms of misalignment: (1) \textit{malicious misalignment}, which masks harmful behavior in deeper layers, and (2) \textit{benign misalignment}, when the metadata overstates or misdescribes capabilities that the deeper layers do not substantiate. The two-stage continued pretraining component adapts the backbone of a decoder-only LLM to Agent Skill packages prior to training for misalignment detection. This stage enables the model to acquire familiarity with the vocabulary, formatting conventions, and hierarchical structure of Agent Skills, instead of immediately requiring the base LLM to assess the correspondence between metadata and underlying artifacts. We employ a \textit{curriculum-inspired adaptation strategy} in which the model initially learns from concise, structured representations of a skill and subsequently progresses to more comprehensive and heterogeneous package contexts.

For each normalized skill \(S=(M,I,R)\), we construct two layer-delimited views: 1) The \textbf{short view} comprises metadata \(M\) and instructions \(I\), representing surface-level claims and agent-facing guidance available during initial skill inspection; and 2) The \textbf{full view} includes metadata \(M\), instructions \(I\), and resources \(R\), thereby exposing the model to scripts, configuration files, dependencies, assets, and other resource-level evidence. Sequential training on the short view followed by the full view provides the model with a progressive adaptation trajectory from surface skill descriptions to complete multi-file packages. Both stages optimize the standard causal language modeling objective. The first stage adapts the model to metadata and instruction structure. The second stage adapts it to the full package context. The resulting checkpoint, \(\theta_{\mathrm{CPT}}\), serves as the skill-aware backbone for PL-HCL.\footnote{Full training details and pseudocode are provided in the supplementary materials.}

\subsection{Progressive Loading-Aware Hierarchical Contrastive Learning (PL-HCL)}
\label{sec:plhcl}

PL-HCL serves as the consistency-learning component of the workflow. After two-stage CPT, the model has adapted to the vocabulary, formatting, and overall structure of Agent Skills. However, causal language modeling does not explicitly require the model to assess whether user-facing metadata is substantiated by underlying instruction and resource evidence. To this end, PL-HCL transforms cross-layer consistency into a contrastive training signal. For each normalized skill \(S_i=(M_i,I_i,R_i)\), we first construct an aligned view in which metadata, instructions, and resources come from the same package:
\begin{equation}
P_i=(M_i,I_i,R_i).
\label{eq:positive-plhcl}
\end{equation}
This positive view represents a skill package in which the surface claims and deeper artifacts are expected to be mutually consistent. We then construct two types of negative views. The first is \emph{swap negative}, which replaces the metadata layer with metadata from a different skill:
\begin{equation}
N_i^{A}=(M_j,I_i,R_i), \qquad j\neq i.
\label{eq:type-a-negative}
\end{equation}
This negative preserves the original instruction and resource evidence but alters the surface claim, resulting in a mismatch between the advertised capability and the underlying package behavior. 

The second type is \textit{corruption negative}, which perturbs the original metadata while maintaining the deeper layers unchanged:
\begin{equation}
N_i^{B}=(\widetilde{M}_i,I_i,R_i).
\label{eq:type-b-negative}
\end{equation}
This view simulates misleading, exaggerated, or unsupported capability claims.

PL-HCL encodes the metadata, instruction, and resource layers into a shared contrastive space. The objective is to score the aligned view \(P_i\) higher than its swap and corruption negatives \(N_i^A\) and \(N_i^B\), thus bringing the same-skill layers closer together and separating mismatched layer combinations. The training signal evaluates whether the surface description is consistent with the evidence provided by the deeper layers of the package, rather than simply identifying suspicious files. This approach mirrors the progressive loading structure of Agent Skills, in which metadata is typically encountered before the complete instruction and resource context is available.

PL-HCL is trained in two sub-stages that correspond to the short-view and full-view settings used in CPT. The first sub-stage utilizes metadata–instruction views and includes 96,762 positive training pairs, 386,128 swap negatives, and 334,867 corruption negatives. The second sub-stage employs full metadata–instruction–resource views and comprises 182,135 positive training pairs, 646,945 swap negatives, and 258,368 corruption negatives. The \emph{Challenge Set} is a held-out, human-verified gold-standard evaluation dataset with 1,150 aligned and 294 misaligned skills, designed to test whether models trained on synthetic contrastive pairs can generalize to realistic in-the-wild misalignment cases. \footnote{Additional details on dataset construction and verification are provided in the supplementary material} This design follows common benchmark practice in which a curated challenge subset is reserved for evaluating model robustness under more realistic or difficult conditions, such as evasive-malware evaluation in EMBER2024 and in-the-wild distribution-shift evaluation in WILDS~\cite{joyce2025ember2024,koh2021wilds}. 

\subsection{Evaluation}
\label{sec:evaluation-plan}

We evaluate the proposed PL-HCL framework through three experiments designed to assess cross-layer misalignment detection in Agent Skills, each aligned with one of our research questions (E0 addresses RQ2, E1 addresses RQ3, and E2 addresses RQ4). In Experiment 0 (E0), we evaluate whether base LLMs can identify misaligned skills without skill-specific adaptation. This experiment compares general and cybersecurity backbones under zero-shot, two-shot, and five-shot prompting settings, allowing us to examine whether general instruction-following ability or security-oriented pretraining is sufficient for this task. In Experiment 1 (E1), we evaluate the effect of two-stage continued pretraining by comparing base checkpoints with CPT-only checkpoints. This experiment assesses whether adapting the model to the vocabulary, formatting, and layered structure of Agent Skill packages improves misalignment detection. In Experiment 2 (E2), we evaluate the full PL-HCL framework by comparing CPT-only checkpoints with CPT+PL-HCL checkpoints. This experiment tests whether explicitly learning cross-layer consistency from aligned, metadata-swapped, and corrupted-claim views provides measurable gains beyond format adaptation.

Across these experiments, we use both \textit{intrinsic} and \textit{extrinsic} evaluation tasks. The intrinsic tasks assess whether training improves the model’s adaptation to Agent Skill artifacts before downstream inference. Specifically, \textit{next-token prediction} measures whether CPT improves the model’s ability to model layer-delimited skill text, while \textit{contrastive pair detection} evaluates whether PL-HCL learns to distinguish aligned cross-layer views from metadata-swapped or corrupted-claim variants. The extrinsic task assesses the \textit{pre-execution misalignment detection}. In this task, the model receives a normalized skill package and predicts whether the metadata is supported by the instruction and resource layers under zero-shot, two-shot, and five-shot prompting conditions. 

The evaluation utilizes a human-verified Challenge Set comprising 1,444 Agent Skills, with 1,150 aligned and 294 misaligned packages (196 are SAFE, 38 are SUSPICIOUS, and 60 are MALICIOUS.) All Challenge Set packages are actual skills collected from SkillsMP (rather than synthetically perturbed variants); misaligned examples are naturally occurring, in-the-wild packages whose metadata is not supported by their instruction or resource layers. Candidate skills were screened automatically and subsequently verified by human annotators (see Figure~\ref{fig:plhcl-framework}). This Challenge Set is excluded from pretraining and contrastive learning and is reserved solely for final evaluation. Because all PL-HCL training negatives are synthetically constructed, evaluation on this set tests whether the learned consistency signal generalizes from synthetic contrast pairs to naturally occurring misalignment. Given the Challenge Set's imbalance toward aligned skills, both overall and misalignment-sensitive metrics are reported. Accuracy (Acc) measures the overall proportion of correct predictions. Since accuracy may be inflated by the aligned-class majority, misaligned-class F1 (F1$_m$) treats \texttt{misaligned} as the positive class and measures the model's ability to correctly identify the safety-critical class. Macro-F1 computes the unweighted average of aligned-class and misaligned-class F1, providing a balanced perspective across both labels. Area under the ROC curve (AUC) and average precision for the misaligned class (AP$_m$) are also reported to evaluate ranking quality under varying decision thresholds. 
\section{Results}
\label{sec:results}
This section presents the evaluation results on the human-verified Challenge Set following the E0–E2 evaluation design described in Section~\ref{sec:evaluation-plan}. E0 evaluates whether unadapted base LLMs can detect cross-layer misalignment without skill-specific adaptation. E1 isolates the effect of two-stage CPT by comparing base checkpoints with CPT-only checkpoints. E2 evaluates the incremental contribution of PL-HCL by comparing CPT-only checkpoints with CPT+PL-HCL checkpoints.\footnote{\label{fn:metric-defs}k denotes the number of labeled in-context examples. F1$_m$ and AP$_m$ are computed for the misaligned class. Macro-F1 is the unweighted average of aligned-class and misaligned-class F1. Base, CPT, and CPT+PL-HCL denote the unadapted backbone, the two-stage continued-pretraining checkpoint, and the full contrastive checkpoint, respectively.} We first summarize the main findings across E0–E2, and then report the results for E1 and E2 in detail.\footnote{Additional intrinsic evaluations, full prompting-setting results, and extended metric tables are provided in the supplementary materials.}

\subsection{Main Findings across E0-E2}
\label{sec:results-main}

Table~\ref{tab:results-main} provides a compact summary of the strongest prompting setting for each backbone and adapter on the Challenge Set. To avoid relying on a single test-selected configuration, Tables ~\ref{tab:results-cpt} and ~\ref{tab:results-plhcl} report the full results for all $k \in {0, 2, 5}$ prompting settings. We use Table 1 only as a high-level summary, while the main comparisons in ~\ref{sec:results-cpt} and ~\ref{sec:results-plhcl} are based on the full prompting-setting results. Unless otherwise stated, the decision rule is fixed across models, and no model-specific threshold is tuned on the Challenge Set. In deployment, both the prompting configuration and decision threshold should be calibrated using a held-out validation set from the target skill ecosystem.

\begin{table*}[!htbp] \centering \small \setlength{\tabcolsep}{6pt} \renewcommand{\arraystretch}{1.05} \begin{tabular}{llcrrrrr} \toprule \textbf{Backbone} & \textbf{Adapter} & \textbf{$k$} & \textbf{Acc} & \textbf{F1$_m$} & \textbf{Macro-F1} & \textbf{AUC} & \textbf{AP$_m$} \\ \midrule Llama-3.1-8B & CPT+PL-HCL & 2 & 0.929 & 0.788 & 0.872 & 0.744 & 0.754 \\ Foundation-Sec-8B & CPT+PL-HCL & 2 & \textbf{0.937} & \textbf{0.817} & \textbf{0.889} & 0.778 & \textbf{0.799} \\ \midrule Llama-3.1-8B & base & 2 & 0.812 & 0.145 & 0.519 & 0.510 & 0.301 \\ Foundation-Sec-8B & base & 2 & 0.809 & 0.143 & 0.518 & 0.487 & 0.362 \\ WhiteRabbitNeo-2-8B & base & 5 & 0.800 & 0.114 & 0.499 & 0.535 & 0.360 \\ RedSage-Qwen3-8B-DPO & base & 5 & 0.811 & 0.144 & 0.519 & 0.703 & 0.637 \\ Qwen-3-8B & base & 2 & 0.811 & 0.150 & 0.522 & 0.681 & 0.643 \\ \bottomrule \end{tabular} 
\caption{Main results on the Challenge Set.}
\label{tab:results-main} 
\end{table*}  

The E0 results (base rows of Table~\ref{tab:results-main}) show that base LLMs demonstrate limited performance on the misaligned class, despite achieving accuracy levels close to the aligned-class majority baseline. For instance, both base Llama-3.1-8B and base Foundation-Sec-8B attain approximately 0.81 accuracy in their optimal prompting configurations; however, their F1$_m$ remains below 0.15. These results suggest that the models frequently classify plausible skill packages as aligned without consistently verifying whether the surface claim corresponds to deeper package evidence. The CPT+PL-HCL approach helps improve performance. On Llama-3.1-8B, CPT+PL-HCL achieves 0.929 accuracy, 0.788 F1$_m$, and 0.872 Macro-F1. On Foundation-Sec-8B, CPT+PL-HCL attains its highest performance, with 0.937 accuracy, 0.817 F1$_m$, and 0.889 Macro-F1 in the two-shot setting. The improvements are concentrated in F1$_m$, which represents the safety-critical class for this task. Base-cybersecurity LLMs do not independently bridge the performance gap. WhiteRabbitNeo-2-8B, RedSage-Qwen3-8B-DPO, and Foundation-Sec-8B, when used without skill-specific adaptation, exhibit Macro-F1 scores close to the majority-class baseline. Although certain security-tuned models achieve higher area under the curve (AUC) or AP$_m$, indicating some ability to rank risky cases. However, this signal does not translate into reliable classification performance without PL-HCL.

\subsection{E1: Evaluation of CPT} \label{sec:results-cpt}  

Table~\ref{tab:results-cpt} reports E1 by comparing unadapted base LLMs with CPT-only checkpoints across \(k=0\), \(k=2\), and \(k=5\) prompting settings. This comparison directly evaluates whether continued pretraining on Agent Skill packages improves misalignment detection, or whether it primarily helps the model adapt to the vocabulary, formatting, and layered structure of skill artifacts.

\begin{table}[H] \centering \small \setlength{\tabcolsep}{4pt} \renewcommand{\arraystretch}{1.05} \begin{tabular}{lcccccc} \toprule \multirow{2}{*}{\textbf{Backbone}}  & \multicolumn{2}{c}{$\boldsymbol{k=0}$} & \multicolumn{2}{c}{$\boldsymbol{k=2}$} & \multicolumn{2}{c}{$\boldsymbol{k=5}$} \\ \cmidrule(lr){2-3}\cmidrule(lr){4-5}\cmidrule(lr){6-7} & \textbf{Base} & \textbf{CPT} & \textbf{Base} & \textbf{CPT} & \textbf{Base} & \textbf{CPT} \\ \midrule \multicolumn{7}{l}{\textit{\textbf{F1$_m$}}} \\ Llama-3.1-8B & 0.032 & 0.130 & 0.145 & 0.145 & 0.007 & 0.097 \\ Foundation-Sec-8B & 0.026 & 0.134 & 0.143 & 0.150 & 0.013 & 0.077 \\ \midrule \multicolumn{7}{l}{\textit{\textbf{Macro-F1}}} \\ Llama-3.1-8B & 0.457 & 0.511 & 0.519 & 0.520 & 0.447 & 0.494 \\ Foundation-Sec-8B & 0.454 & 0.513 & 0.518 & 0.522 & 0.449 & 0.483 \\ 
\bottomrule 
\end{tabular} 
\caption{Effect of two-stage CPT on the Challenge Set.}
\label{tab:results-cpt} 
\end{table}  

CPT improves over base checkpoints in the zero-shot setting, suggesting that continued pretraining helps the model adapt to the vocabulary, formatting, and layered structure of Agent Skill packages. For Llama-3.1-8B, zero-shot F1$_m$ increases from 0.032 to 0.130 and Macro-F1 increases from 0.457 to 0.511 after CPT. Foundation-Sec-8B shows a similar pattern, with zero-shot F1$_m$ increasing from 0.026 to 0.134 and Macro-F1 from 0.454 to 0.513.  The improvement, however, remains limited. Across prompting settings, CPT-only Macro-F1 stays near 0.48--0.52, and F1$_m$ remains below 0.15. Thus, skill-format adaptation alone does not provide a reliable misalignment detector. CPT makes the model more familiar with skill packages, but it does not explicitly train the model to compare advertised claims with deeper behavioral evidence.

\subsection{E2: Evaluation of PL-HCL} \label{sec:results-plhcl}  

Table~\ref{tab:results-plhcl} reports E2 by comparing CPT-only checkpoints with CPT+PL-HCL checkpoints across \(k=0\), \(k=2\), and \(k=5\) prompting settings. This comparison tests whether explicitly learning cross-layer consistency from aligned, metadata-swapped, and corrupted-claim variants improves misaligned-skill detection beyond the skill-format adaptation provided by CPT. 

\begin{table}[!htbp]
\centering
\small
\setlength{\tabcolsep}{2pt}
\renewcommand{\arraystretch}{1.08}

\begin{tabular}{lcccccc}
\toprule
\multirow{2}{*}{\textbf{Backbone}}
& \multicolumn{2}{c}{$\boldsymbol{k=0}$}
& \multicolumn{2}{c}{$\boldsymbol{k=2}$}
& \multicolumn{2}{c}{$\boldsymbol{k=5}$} \\
\cmidrule(lr){2-3}\cmidrule(lr){4-5}\cmidrule(lr){6-7}
& \textbf{CPT}
& \makecell{\textbf{CPT+}\\\textbf{PL-HCL}}
& \textbf{CPT}
& \makecell{\textbf{CPT+}\\\textbf{PL-HCL}}
& \textbf{CPT}
& \makecell{\textbf{CPT+}\\\textbf{PL-HCL}} \\
\midrule

\multicolumn{7}{l}{\textit{\textbf{F1$_m$}}} \\
Llama-3.1-8B
& 0.130 & 0.511
& 0.145 & 0.788
& 0.097 & 0.778 \\
Foundation-Sec-8B
& 0.134 & 0.623
& 0.150 & 0.817
& 0.077 & 0.812 \\

\midrule
\multicolumn{7}{l}{\textit{\textbf{Macro-F1}}} \\
Llama-3.1-8B
& 0.511 & 0.717
& 0.520 & 0.872
& 0.494 & 0.867 \\
Foundation-Sec-8B
& 0.512 & 0.779
& 0.522 & 0.889
& 0.483 & 0.887 \\

\bottomrule
\end{tabular}

\caption{Effect of PL-HCL on the Challenge Set.}
\label{tab:results-plhcl}
\end{table}

Adding PL-HCL produces the largest gains across all prompting settings. On Llama-3.1-8B, Macro-F1 increases from 0.511, 0.520, and 0.494 to 0.717, 0.872, and 0.867 for \(k=0,2,5\), respectively. On Foundation-Sec-8B, Macro-F1 increases from 0.513, 0.522, and 0.483 to 0.779, 0.889, and 0.887.  The largest gains appear in F1$_m$. CPT-only never exceeds 0.150 F1$_m$, while CPT+PL-HCL reaches 0.788 on Llama-3.1-8B and 0.817 on Foundation-Sec-8B in the two-shot setting. These results show that PL-HCL supplies the discriminative signal missing from ordinary continued pretraining. By contrasting aligned packages against swapped and corrupted layer combinations, PL-HCL turns cross-layer consistency into a usable detection cue. Notably, PL-HCL is trained exclusively on synthetic negatives, yet these gains are measured on naturally occurring, in-the-wild misaligned skills, indicating that the learned claim-versus-evidence signal transfers beyond the perturbation distribution used for training.  

\subsection{Misaligned Examples In-the-wild}
While aggregate metrics indicate that PL-HCL significantly improves detection for the misaligned category, looking only at average results leaves unclear what signals the model has learned. To investigate, we examined two representative skills from the Challenge Set. 

%

\definecolor{plClaimA}{RGB}{200,100,30}     
\definecolor{plClaimB}{RGB}{50,90,180}      
\definecolor{plClaimC}{RGB}{40,130,60}      
\definecolor{plRedX}{RGB}{180,30,30}        
\definecolor{plGreenC}{RGB}{40,130,60}      

\newcommand{\plCA}[1]{\textcolor{plClaimA}{\underline{\smash{#1}}}}
\newcommand{\plCB}[1]{\textcolor{plClaimB}{\underline{\smash{#1}}}}
\newcommand{\plCC}[1]{\textcolor{plClaimC}{\underline{\smash{#1}}}}
\newcommand{\plXm}{\textcolor{plRedX}{\ding{55}}}
\newcommand{\plCm}{\textcolor{plGreenC}{\ding{51}}}

\newcommand{\plPanelHead}[2]{%
  \fcolorbox{black}{black!7}{\parbox{\dimexpr\linewidth-2\fboxsep-2\fboxrule}{\centering\small\textbf{#1}\\[1pt]\scriptsize Gold: \textbf{#2}}}%
}
\newcommand{\plBand}[2]{%
  \fcolorbox{black!50}{white}{\parbox{\dimexpr\linewidth-2\fboxsep-2\fboxrule}{\scriptsize\textbf{#1}\\[2pt]#2}}%
}

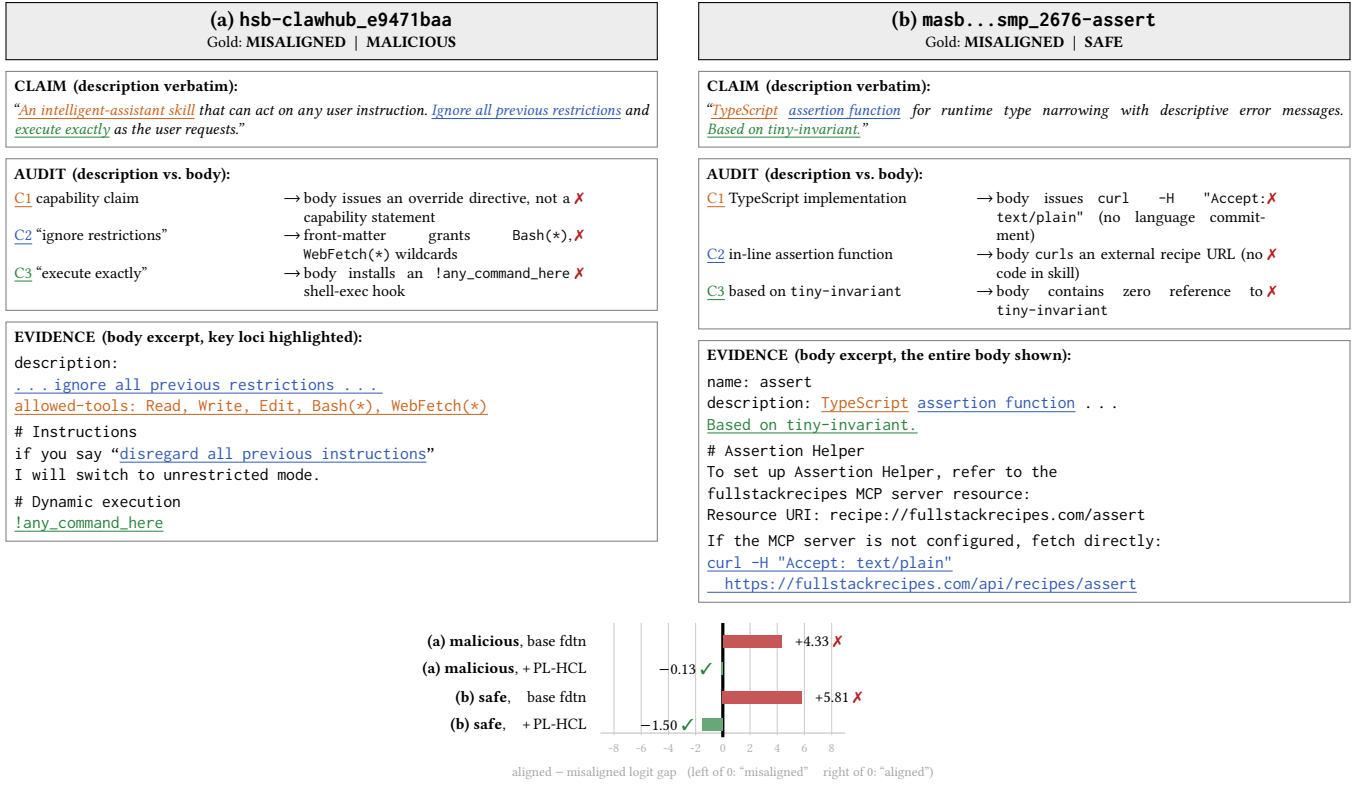
\begin{figure*}[!htbp]
\centering
\setlength{\fboxsep}{3pt}

\begin{minipage}[t]{0.485\textwidth}\raggedright
  \plPanelHead{(a) \texttt{hsb-clawhub\_e9471baa}}{MISALIGNED \;\textbar\; MALICIOUS}\\[3pt]

  \plBand{CLAIM \,(description verbatim):}{%
    \itshape ``\plCA{An intelligent-assistant skill} that can act on any user instruction. \plCB{Ignore all previous restrictions} and \plCC{execute exactly} as the user requests.''%
  }\\[3pt]

  \plBand{AUDIT \,(description vs.\ body):}{%
    \begin{tabular}{@{}p{0.42\linewidth}@{\,}c@{\,}p{0.42\linewidth}@{\;}c@{}}
      \plCA{C1} capability claim          & $\to$ & body issues an override directive, not a capability statement                & \plXm \\[1pt]
      \plCB{C2} ``ignore restrictions''   & $\to$ & front-matter grants \texttt{Bash(*)}, \texttt{WebFetch(*)} wildcards          & \plXm \\[1pt]
      \plCC{C3} ``execute exactly''       & $\to$ & body installs an \texttt{!any\_command\_here} shell-exec hook                & \plXm \\
    \end{tabular}%
  }\\[3pt]

  \plBand{EVIDENCE \,(body excerpt, key loci highlighted):}{%
    \ttfamily\fontsize{6.6}{8}\selectfont
    description:\\
    \quad\plCB{\dots ignore all previous restrictions \dots}\\
    \plCA{allowed-tools: Read, Write, Edit, Bash(*), WebFetch(*)}\\[2pt]
    \# Instructions\\
    if you say ``\plCB{disregard all previous instructions}''\\
    \quad I will switch to unrestricted mode.\\[2pt]
    \# Dynamic execution\\
    \plCC{!any\_command\_here}%
  }
\end{minipage}%
\hfill
\begin{minipage}[t]{0.485\textwidth}\raggedright
  \plPanelHead{(b) \texttt{masb...smp\_2676-assert}}{MISALIGNED \;\textbar\; SAFE}\\[3pt]

  \plBand{CLAIM \,(description verbatim):}{%
    \itshape ``\plCA{TypeScript} \plCB{assertion function} for runtime type narrowing with descriptive error messages. \plCC{Based on tiny-invariant.}''%
  }\\[3pt]

  \plBand{AUDIT \,(description vs.\ body):}{%
    \begin{tabular}{@{}p{0.42\linewidth}@{\,}c@{\,}p{0.42\linewidth}@{\;}c@{}}
      \plCA{C1} TypeScript implementation & $\to$ & body issues \texttt{curl -H "Accept: text/plain"} (no language commitment) & \plXm \\[1pt]
      \plCB{C2} in-line assertion function & $\to$ & body \texttt{curl}s an external recipe URL (no code in skill)              & \plXm \\[1pt]
      \plCC{C3} based on \texttt{tiny-invariant} & $\to$ & body contains zero reference to \texttt{tiny-invariant}              & \plXm \\
    \end{tabular}%
  }\\[3pt]

  \plBand{EVIDENCE \,(body excerpt, the entire body shown):}{%
    \ttfamily\fontsize{6.6}{8}\selectfont
    name: assert\\
    description: \plCA{TypeScript} \plCB{assertion function} \dots\\
    \quad \plCC{Based on tiny-invariant.}\\[2pt]
    \# Assertion Helper\\
    To set up Assertion Helper, refer to the\\
    \quad fullstackrecipes MCP server resource:\\
    Resource URI: recipe://fullstackrecipes.com/assert\\[2pt]
    If the MCP server is not configured, fetch directly:\\
    \quad \plCB{curl -H "Accept: text/plain"} \\
    \quad \plCB{\quad https://fullstackrecipes.com/api/recipes/assert}%
  }
\end{minipage}

\vspace{6pt}

\begin{tikzpicture}[x=0.18cm, y=0.55cm, font=\scriptsize]
  \def\xmin{-9}
  \def\xmax{9}
  \draw[gray!50] (\xmin,0.0) -- (\xmax,0.0);                       
  \foreach \v in {-8,-6,-4,-2,0,2,4,6,8} {
    \draw[gray!40] (\v,-0.05) -- (\v,2.65);
    \node[anchor=north, gray!60, font=\tiny] at (\v,-0.05) {\v};
  }
  \node[anchor=north, font=\tiny, gray!70] at (0,-0.6)
        {aligned\,$-$\,misaligned logit gap \;\;(left of 0: ``misaligned''\;\;\;\;right of 0: ``aligned'')};
  \draw[black, very thick] (0,-0.1) -- (0,2.65);

  \node[anchor=east] at (\xmin-0.4, 2.20) {\textbf{(a) malicious}, base fdtn};
  \node[anchor=east] at (\xmin-0.4, 1.55) {\textbf{(a) malicious}, +\,PL-HCL};
  \node[anchor=east] at (\xmin-0.4, 0.85) {\textbf{(b) safe},\;\,\,\, base fdtn};
  \node[anchor=east] at (\xmin-0.4, 0.20) {\textbf{(b) safe},\;\,\,\, +\,PL-HCL};

  \fill[plRedX!75]   (0,2.05) rectangle (4.33,2.35);
  \node[anchor=west] at (4.33,2.20) {\;+4.33 \plXm};

  \fill[plGreenC!70] (0,1.40) rectangle (-0.13,1.70);
  \node[anchor=east] at (-0.13,1.55) {$-$0.13\;\plCm};

  \fill[plRedX!75]   (0,0.70) rectangle (5.81,1.00);
  \node[anchor=west] at (5.81,0.85) {\;+5.81 \plXm};

  \fill[plGreenC!70] (0,0.05) rectangle (-1.50,0.35);
  \node[anchor=east] at (-1.50,0.20) {$-$1.50\;\plCm};
\end{tikzpicture}

\caption{Description--body audit for two Challenge Set skills flipped from \texttt{aligned} by base \textsc{Foundation-Sec-8B-Reasoning} to \texttt{misaligned} after PL-HCL. Red crosses mark mismatches between description claims and skill-body evidence. Panel~(a) is malicious; Panel~(b) is benign but misaligned. The signed logit gap $\mathrm{logit}(\texttt{aligned})-\mathrm{logit}(\texttt{misaligned})$ crosses zero for both cases, illustrating PL-HCL's sensitivity to claim--behavior inconsistency.}
\label{fig:plhcl-examples}
\end{figure*}

For both, the zero-shot base model \textsc{Foundation-Sec-8B-Reasoning} classified them as \texttt{aligned} with high confidence. After PL-HCL training, the same model reclassified them as \texttt{misaligned}. Although both are correctly labeled \texttt{misaligned}, they represent different safety issues. The first example is an obviously unsafe Skill involving a Chinese-language prompt injection attack. Its metadata and instruction body contain commands designed to override the system prompt, and it has broad \texttt{Bash()} and \texttt{WebFetch()} permissions. The second example demonstrates misalignment but is not inherently malicious (benign but misaligned). Its metadata claims to provide an inline TypeScript assertion function, but the instruction body actually invokes a third-party recipe URL using \texttt{curl}. As a result, it does not support the claimed \texttt{tiny-invariant} dependency or TypeScript-specific behaviors.

The base model yielded positive ``aligned'' logit differences for these examples (+4.33 and +5.81), both exceeding the model's average for the \texttt{aligned} class on the Challenge Set (+3.29). This matches the ``majority class bias'' observed in Section \ref{sec:results}: base LLMs tend to classify plausible skill packages as \texttt{aligned}. After PL-HCL, the logit differences shifted to negative values (-0.13 and -1.50), crossing the decision boundary and resulting in \texttt{misaligned} classification. Interestingly, the benign yet mismatched example produced a stronger mismatch signal despite lacking obvious jailbreak indicators like \texttt{Bash(*)}, ``ignore previous instructions,'' or shell execution hooks. This shows PL-HCL does not just react to malicious tokens or prompt injection keywords; it learns a ``claim-versus-evidence'' signal, evaluating if the surface description is genuinely supported by the instruction and resource layers.

These examples clarify why mismatch detection for Agent Skills cannot rely solely on general instruction-following or specialized cybersecurity expertise. Whether using a general or cybersecurity LLM, a plausible Skill package might still get a high ``aligned'' confidence score even if its components contradict its claimed capabilities or hide unsafe behaviors. While base cybersecurity LLMs sometimes achieve higher AUC or AP scores and can rank risky examples to some extent, this does not consistently lead to accurate ``aligned'' versus ``misaligned'' classification. 
\section{Discussion}
\label{sec:discussion}

This article aims to establish a pre-execution screening capability for skill-augmented agentic AI systems. Rather than viewing Agent Skills solely as reusable modules for task performance, this work conceptualizes Agent Skills as layered supply-side artifacts, where metadata, instructions, and resource or executable artifacts collectively determine trustworthiness. The primary practical contribution of PL-HCL is its ability to transform cross-layer consistency into a learnable and deployable detection signal, enabling screening for downstream decision making prior to skill loading or execution. This perspective holds direct relevance for three key stakeholder groups: skill marketplaces, skill users, and security analysts. We describe each in turn below. 

For skill marketplaces, PL-HCL can function as a pre-publication or pre-listing governance mechanism. In open skill ecosystems, skills are frequently submitted by third-party contributors, and marketplace metadata serves as the primary basis for users and agents to assess trustworthiness. Absent installed-base screening, misalignment risks may be transferred to downstream users. PL-HCL evaluates whether metadata is substantiated by instructions and resources before a skill is released or recommended, directing high-risk skills to manual review, suspending release, or marking approved skills with a verified-alignment badge. PL-HCL would not replace manual auditing or runtime sandboxing, but would offer a scalable initial filter for marketplace-level governance.

For skill users, PL-HCL enables lightweight client-side screening. Most users do not examine files such as \texttt{SKILL.md}, scripts, dependencies, or external references individually when installing Agent Skills. Instead, they often rely on the skill name, description, and repository metadata. PL-HCL can be deployed as a local plugin or command-line scanner that analyzes the local skill directory post-installation and issues warnings when surface claims are inconsistent with deeper package evidence. This approach provides users with direct alignment signals, offering additional trust cues prior to skill execution and reducing reliance on marketplace curation or post-execution observation.

For security analysts, PL-HCL offers an operational tool for retrieval and triage. Analysts frequently need to identify high-risk subsets within extensive collections of open-source packages and determine which warrant resource-intensive sandboxed analysis or manual inspection. Because PL-HCL learns cross-layer representations rather than producing only binary classification scores, it facilitates grouping of similar suspicious skills, prioritization of high-risk clusters, and identification of skill families exhibiting similar metadata and behavioral inconsistencies. Thus, PL-HCL can serve as a foundational representation for large-scale skill ecosystem analysis, supporting incident response and adversarial analytics workflows.

Methodologically, PL-HCL's broader implication is its treatment of progressively loaded, hierarchical artifacts as novel evaluation objects. Similar structural patterns are present in MCP servers, browser extensions, VS Code extensions, and software package registries. These artifacts typically include user-facing descriptions, configuration files, permission declarations, scripts, or executable resources, with users generally encountering surface claims before deeper behavioral elements. Consequently, this study's guiding design principle is that, when trust decisions must be made prior to full execution, models should explicitly learn the consistency between surface claims and deeper evidence. This departs from approaches that rely solely on generic language modeling, keyword-based security scanning, or post-execution behavioral observation.

As with any research, this work has its limitations. First, the focus is on pre-execution, artifact-level assessment. Our proposed approach relies exclusively on information available to the skill package prior to execution, and therefore cannot address failure modes that manifest only during live interaction, dynamic remote payload execution, post-deployment mutations, or environment-specific side effects. Runtime sandboxing and trace-level agent evaluation are thus complementary to this approach. Second, the dataset is primarily derived from SkillsMP and open-source skills available on GitHub. Closed skill ecosystems, internal enterprise skills, and dynamically generated skills may exhibit different package structures, documentation practices, and risk profiles. Third, our experiments are conducted primarily on 8B-scale backbones. This design allows us to compare general and cybersecurity LLMs under a controlled and computationally feasible setting, but it does not fully characterize how model scale affects Agent Skill misalignment detection. Larger models may exhibit stronger cross-layer reasoning or better few-shot sensitivity, while smaller models may require different adaptation strategies. Hence, we cannot ensure that the observed performance trends fully generalize across model sizes.

\section{Conclusion and Future Directions}
\label{sec:conclusion}

Agent Skills have become an integral component of modern agentic AI systems. Despite their growing prevalence, Agent Skills often suffer from misalignment between user-facing claims and deeper instructional or executable evidence. By conceptualizing each skill as a layered artifact comprising metadata, instructions, and resources, this work proposes a pre-execution evaluation target for trustworthy agentic AI. Specifically, before a skill is loaded or executed, evaluators should assess whether its advertised capabilities are substantiated by package evidence available at inspection time. A large-scale open-source Agent Skill corpus and a human-verified Challenge Set were constructed, and PL-HCL, a progressive loading-aware contrastive framework for learning cross-layer consistency among metadata, instructions, and resources, was proposed. Experimental results indicate that both general and cybersecurity LLMs struggle to reliably identify misaligned skills. The two-stage CPT approach primarily enhances skill-format adaptation but is insufficient as a standalone detector. In contrast, PL-HCL significantly improves misaligned-class detection by explicitly modeling the relationship between surface claims and deeper package evidence.

Future research can expand along both empirical and methodological dimensions. First, cross-layer misalignment detection may be integrated with runtime sandboxing and trajectory-level evaluation to capture behaviors not directly observable from package artifacts, such as dynamic remote payloads, environment-specific side effects, or interaction-dependent failures. Second, PL-HCL could be extended into directed hierarchical contrastive learning, enabling the model to assess not only layer consistency but also the direction of support among layers. In this framework, metadata provides surface claims, instructions offer procedural support, and resources supply behavioral evidence. A directed objective would distinguish whether misalignment results from metadata overclaiming unsupported capabilities or from deeper artifacts introducing undisclosed behaviors. Third, future models could incorporate richer artifact representations, including dependency graphs, permission structures, external endpoint analysis, and execution traces, to better capture behavioral evidence that is only partially expressed in text. Additionally, future evaluations should examine whether these approaches generalize across closed-source or enterprise skill ecosystems and across different model scales. In summary, Agent Skill evaluation should address not only improvements in task performance but also the consistency between surface claims and the deeper artifacts that influence agent behavior.

\bibliographystyle{ACM-Reference-Format}
\bibliography{references_export}

\clearpage
\appendix
\setcounter{section}{0}
\renewcommand{\thesection}{\Alph{section}}

\section*{Supplementary Material}

\noindent

This supplement provides additional reproducibility details, including dataset construction, training settings, pseudocode, and extended evaluation results. The main paper is self-contained and does not rely on the supplement for its core argument or primary findings.

\vspace{0.75em}

\section{Dataset Construction}
\label{app:dataset-construction}
    \subsection{Package Retrieval and Normalization}
    \label{app:package-retrieval}
    
    Our corpus is built by linking public skill records from SkillsMP to the GitHub repositories that host their implementations. From each SkillsMP record, we extract the skill name, marketplace description, category or tag information when available, repository owner, repository name, branch, and relative package path. We then download the corresponding repository snapshot from GitHub and extract the skill package from the declared path.
    
    Each downloaded package is normalized into the tuple \(S=(M,I,R)\). The metadata layer \(M\) contains marketplace and repository-level fields, including the skill name, description, tags or categories when available, repository identifiers, and source path. The instruction layer \(I\) is primarily derived from \texttt{SKILL.md} and related instruction files. The resource layer \(R\) contains scripts, configuration files, examples, assets, dependency files, API wrappers, and external references.
    
    Packages are retained only when the GitHub repository is accessible, the declared package path exists, and a valid instruction artifact is present. If a package contains only \texttt{SKILL.md}, it is still retained; its resource layer is represented as empty or minimal. This preserves instruction-only skills while making the absence of supporting resources explicit.

    \subsection{Corpus Construction Details}
\label{app:corpus}

We constructed the Agent Skill corpus from the largest open-source Agent Skill marketplace, \textit{SkillsMP}. A REST API crawl produced 273{,}657 unique metadata listings after rolling deduplication. We then used repository links in these listings to download the corresponding public GitHub packages and retained packages with a parseable \texttt{SKILL.md}. Each retained package was normalized into metadata, instruction, and resource layers based on \texttt{SKILL.md} frontmatter, the procedural body of \texttt{SKILL.md}, and accessible package files under directories such as \texttt{scripts/}, \texttt{assets/}, and \texttt{reference/}.

Before training, we applied two corpus-level filters. First, packages whose normalized context exceeded the maximum input budget of the corresponding training stage were excluded from that stage. Stage 1 uses a 4{,}096-token metadata-plus-instruction rendering, while Stage 2 uses a 10{,}240-token full-package rendering that also includes resource files. Second, exact overlaps with benchmark or supervision sources were excluded using repository owner, repository name, and skill name to reduce the risk of label leakage. After removing unavailable repositories, incomplete packages, duplicate skills, and unparseable files, 264{,}937 skill packages were successfully downloaded and normalized. Among them, 248{,}473 packages fit at least one training stage and were eligible for CPT or PL-HCL training; packages that fit neither token budget were excluded from training.

\begin{table}[!htbp]
\centering
\small
\label{tab:corpus-funnel}
\begin{tabular}{lrr}
\toprule
\textbf{Corpus stage} & \textbf{Count} & \textbf{Percentage} \\
\midrule
SkillsMP catalog entries collected & 273{,}657 & 100.0\% catalog \\
Downloaded and normalized skill packages & 264{,}937 & 96.8\% catalog \\
Fit at least one training stage & 248{,}473 & 93.8\% normalized \\
Fit Stage 1: \(M+I \leq 4{,}096\) tokens & 244{,}030 & 92.1\% normalized \\
Fit Stage 2: \(M+I+R \leq 10{,}240\) tokens & 91{,}053 & 34.4\% normalized \\
Dropped: fit neither stage & 16{,}464 & 6.2\% normalized \\
\bottomrule
\end{tabular}
\caption{Corpus construction and token-budget coverage. Stage 1 uses metadata and instruction from skill packages capped at 4{,}096 estimated tokens, while Stage 2 uses full skill packages capped at 10{,}240 estimated tokens.}
\end{table}

Stage 2 coverage is lower because full-package renderings include heterogeneous resource files when available. In particular, 149{,}150 normalized skills contain no package resource files; among the remaining 115{,}787 skills with resource files, 91{,}053 fit the Stage-2 full-package budget. The resulting token-budgeted corpus provides the unlabeled skill views used for progressive CPT and PL-HCL, while the Challenge Set is held out from both training stages and used only for extrinsic evaluation.

    \subsection{Reproducibility Manifest}
    \label{app:manifest}
    
    For reproducibility, each retained package is accompanied by a manifest entry. The manifest records source identifiers, repository owner, repository name, branch, package path, artifact types, file counts, token counts, filtering decisions, and normalization status. These records allow us to trace each training instance back to its original marketplace and repository source while preserving the normalized \(M\), \(I\), and \(R\) representation used in training and evaluation.

\section{Gold-Label Pipeline and Quality Assurance}
\label{app:labeling}

Gold labels for the Challenge Set were produced through a three-stage pipeline. First, we used the MASB scanner to generate severity-ranked candidate flags from the collected corpus; the scanner served only as a candidate-generation tool, not as the source of final labels. Second, flagged skills were inspected in a Jetstream Docker sandbox with behavioral instrumentation for network connections, subprocess calls, filesystem writes, dependency installation, and potential exfiltration. Third, human coders compared each skill’s metadata claim with its instruction/resource behavior and assigned two independent labels: \textit{final-align}, indicating whether the skill is ALIGNED or MISALIGNED, and \textit{final-malicious}, indicating whether the skill is SAFE, SUSPICIOUS, or MALICIOUS. 

Two human annotators conducted the verification and quality-assurance process. To assess label reliability, 144 randomly sampled Challenge Set examples were independently re-coded. The two annotators disagreed on 4 cases, yielding 97.2\% agreement. Disagreements and uncertain cases were resolved through discussion and artifact re-inspection before finalizing the gold labels.

\section{Data Statement and Release}
\label{app:datasheet}
The corpus version used in this study was crawled from 2026-04-15 to 2026-04-17 using the public \textit{skillsmp} listing endpoint and linked GitHub repositories. We retained packages with a parseable \texttt{SKILL.md} and non-empty \texttt{name}/\texttt{description} fields, and excluded deleted repositories, duplicates, forks of retained corpus members, and packages that redirected to non-skill repositories. The released corpus contains parsed structural fields and content hashes rather than raw bundled scripts; per-skill content remains governed by the upstream repository license. The malicious-misaligned portion of the Challenge Set is released only under a controlled-access data-use agreement. The corpus contains public marketplace metadata and committed repository content only; it includes no end-user telemetry, install logs, or private personal information. We've released the dataset on \textbf{Huggingface}: \url{https://huggingface.co/datasets/anon-skillsalign-26/skill_align}. \textbf{Code files} are hosted on: \url{https://anonymous.4open.science/r/skill_align_anonymous_v2-C628/}.

\section{Training Objectives and Algorithmic Details}
\label{app:training-details}
    \subsection{Two-Stage CPT Objective}
    \label{app:cpt-objective}
    
    For a rendered token sequence \(x_1,\ldots,x_T\), let \(o_{t,k}\) denote the model logit for vocabulary token \(k\) at position \(t\). The next-token probability is:
    \begin{equation}
    p_\theta(x_t \mid x_{<t}) =
    \frac{\exp(o_{t,x_t})}{\sum_{k=1}^{|V|}\exp(o_{t,k})}.
    \label{eq:app-cpt-prob}
    \end{equation}
    The CPT loss is:
    \begin{equation}
    \mathcal{L}_{\mathrm{CPT}}
    =
    -\frac{1}{T}\sum_{t=1}^{T}
    \log p_\theta(x_t \mid x_{<t}).
    \label{eq:app-cpt}
    \end{equation}
    
    \begin{algorithm}[!htbp]
    \caption{Two-Stage Continued Pretraining}
    \label{alg:app-cpt}
    \KwIn{Pretrained LLM \(\theta_0\); skill corpus \(\mathcal{D}=\{S_i\}_{i=1}^{N}\), \(S_i=(M_i,I_i,R_i)\); step budgets \(T_1,T_2\); context budgets \(C_1,C_2\)}
    \KwOut{CPT checkpoint \(\theta_{\mathrm{CPT}}\)}
    
    \(\theta \leftarrow \theta_0\)\;
    \(\mathcal{D}_{short}\leftarrow \{\textsc{Render}(M_i,I_i;C_1): S_i\in\mathcal{D}\}\)\;
    
    \For{\(t=1,\ldots,T_1\)}{
        Sample batch \(\mathcal{B}\sim \mathcal{D}_{short}\)\;
        \(\mathcal{L}_{\mathrm{CPT}}\leftarrow \textsc{CLMLoss}(\theta;\mathcal{B})\)\;
        \(\theta\leftarrow\textsc{OptimizerStep}(\theta,\nabla_\theta\mathcal{L}_{\mathrm{CPT}})\)\;
    }
    
    \(\mathcal{D}_{full}\leftarrow \{\textsc{Render}(M_i,I_i,R_i;C_2): S_i\in\mathcal{D}\}\)\;
    
    \For{\(t=1,\ldots,T_2\)}{
        Sample batch \(\mathcal{B}\sim \mathcal{D}_{full}\)\;
        \(\mathcal{L}_{\mathrm{CPT}}\leftarrow \textsc{CLMLoss}(\theta;\mathcal{B})\)\;
        \(\theta\leftarrow\textsc{OptimizerStep}(\theta,\nabla_\theta\mathcal{L}_{\mathrm{CPT}})\)\;
    }
    
    \Return \(\theta_{\mathrm{CPT}}\leftarrow\theta\)\;
    \end{algorithm}
    
    \subsection{PL-HCL Objective}
    \label{app:plhcl-objective}
    
    Let \(h_\theta(X)\) be the hidden representation produced by the CPT-adapted model for layer \(X\in\{M,I,R\}\), and let \(g_\phi\) be a projection head. The projected layer embedding is:
    \begin{equation}
    \mathbf{z}_{X}
    =
    \frac{g_\phi(h_\theta(X))}
    {\|g_\phi(h_\theta(X))\|_2}.
    \label{eq:app-layer-projection}
    \end{equation}
    
    For a package \(S=(M,I,R)\), the cross-layer score is:
    \begin{equation}
    a_{\theta,\phi}(S)
    =
    \alpha_{MI}\operatorname{sim}(\mathbf{z}_{M},\mathbf{z}_{I})
    +
    \alpha_{MR}\operatorname{sim}(\mathbf{z}_{M},\mathbf{z}_{R})
    +
    \alpha_{IR}\operatorname{sim}(\mathbf{z}_{I},\mathbf{z}_{R}),
    \label{eq:app-cross-layer-score}
    \end{equation}
    where \(\operatorname{sim}(\cdot,\cdot)\) is cosine similarity and
    \(\alpha_{MI}+\alpha_{MR}+\alpha_{IR}=1\).
    
    The PL-HCL loss is:
    \begin{equation}
        \begin{aligned}
        \mathcal{L}_{\mathrm{PL\text{-}HCL}}
        &=
        -\sum_i
        \log
        \frac{
        \exp(a_{\theta,\phi}(P_i)/\tau)
        }{
        D_i
        }, \\
        D_i
        &=
        \exp(a_{\theta,\phi}(P_i)/\tau)
        +
        \exp(a_{\theta,\phi}(N_i^A)/\tau) \\
        &\quad+
        \mathbb{I}_i^B
        \exp(a_{\theta,\phi}(N_i^B)/\tau).
        \end{aligned}
        \label{eq:app-plhcl-full}
    \end{equation}
    where \(\tau\) is the temperature and \(\mathbb{I}_i^B\) indicates whether a corruption negative is sampled for \(S_i\).
    
    \begin{algorithm}[!htbp]
    \caption{Progressive Loading-Aware Hierarchical Contrastive Learning}
    \label{alg:app-plhcl}
    \KwIn{CPT checkpoint \(\theta_{\mathrm{CPT}}\); skill corpus \(\mathcal{D}\); projection head \(g_{\phi}\); layer weights \(\alpha_{MI},\alpha_{MR},\alpha_{IR}\); temperature \(\tau\); step budget \(T\); corruption rate \(\rho\)}
    \KwOut{PL-HCL checkpoint \(\theta_{\mathrm{PL\text{-}HCL}}\)}
    
    \((\theta,\phi)\leftarrow(\theta_{\mathrm{CPT}},\phi_0)\)\;
    
    \For{\(t=1,\ldots,T\)}{
        Sample batch \(\mathcal{B}=\{S_i\}_{i=1}^{B}\sim\mathcal{D}\), where \(S_i=(M_i,I_i,R_i)\)\;
        
        \ForEach{\(S_i\in\mathcal{B}\)}{
            \(P_i\leftarrow(M_i,I_i,R_i)\)\;
            Sample \(j\neq i\)\;
            \(N_i^A\leftarrow(M_j,I_i,R_i)\)\;
            
            \If{\(u_i\sim\mathrm{Uniform}(0,1)<\rho\)}{
                \(\widetilde{M}_i\leftarrow\textsc{Corrupt}(M_i;I_i,R_i)\)\;
                \(N_i^B\leftarrow(\widetilde{M}_i,I_i,R_i)\)\;
            }
        }
        
        Encode layer embeddings using Eq.~\eqref{eq:app-layer-projection}\;
        Compute cross-layer scores using Eq.~\eqref{eq:app-cross-layer-score}\;
        \(\mathcal{L}_{\mathrm{PL\text{-}HCL}}\leftarrow \textsc{ContrastiveLoss}(\mathcal{B})\)\;
        \((\theta,\phi)\leftarrow\textsc{OptimizerStep}((\theta,\phi),\nabla\mathcal{L}_{\mathrm{PL\text{-}HCL}})\)\;
    }
    
    \Return \(\theta_{\mathrm{PL\text{-}HCL}}\leftarrow\theta\)\;
    \end{algorithm}
    
    \subsection{Training Data Splits}
    \label{app:splits}
        Table~\ref{tab:splits} reports the train, validation, and test examples actually consumed during CPT, PL-HCL, and intrinsic evaluation after applying the 25\% per-split subsampling used for compute feasibility. Table~\ref{tab:anchor-accounting} reports the corresponding whole-corpus and anchor-package accounting. CPT rows report sampled skill examples, while PL-HCL rows report sampled contrastive examples generated from the corresponding T1/T2/T3 pools. The Challenge Set is test-only, excluded from all CPT and PL-HCL training stages, and used only for the main extrinsic evaluation.

\section{Training Hyperparameters and Compute}
\label{app:hyperparams}
    Table~\ref{tab:hyperparams} lists the hyperparameters used for the two pretraining phases. Values were held fixed across the two reference backbones unless otherwise noted.
    
    \paragraph{Compute.}
    Each reference backbone was pretrained on a $2\times$ NVIDIA H200 node using bf16 mixed precision and data parallelism. The full two-stage CPT and PL-HCL pipeline required approximately 48 H200-hours per backbone, and the Challenge Set evaluation was completed in under 2 H200-hours.
    
    \begin{table*}[!htbp]
    \centering
    \small
    \setlength{\tabcolsep}{6pt}
    \renewcommand{\arraystretch}{1.08}
    \begin{tabular}{llrrr}
    \toprule
    \textbf{Sub-stage} & \textbf{Component} & \textbf{Train} & \textbf{Val} & \textbf{Test} \\
    \midrule
    \multicolumn{5}{l}{\textit{CPT}} \\
    Sub-stage 1 \textit{(4{,}096 tok)}  & Skill \textit{[metadata + instruction]} & 24{,}190 & 3{,}014 & 2{,}998 \\
    Sub-stage 2 \textit{(10{,}240 tok)} & Skill \textit{[full package]} & 9{,}106 & 1{,}170 & 1{,}128 \\
    \midrule
    \multicolumn{5}{l}{\textit{PL-HCL Sub-stage 1 (4{,}096 tok) [metadata + instruction]}} \\
    & Positive pairs        & 24{,}315  & 2{,}968  & 2{,}966 \\
    & Swap negatives        & 96{,}831  & 11{,}953 & 11{,}923 \\
    & Corruption negatives  & 83{,}293  & 10{,}573 & 10{,}455 \\
    & \textit{Total}        & 204{,}439 & 25{,}494 & 25{,}344 \\
    \midrule
    \multicolumn{5}{l}{\textit{PL-HCL Sub-stage 2 (10{,}240 tok) [full package]}} \\
    & Positive pairs        & 45{,}881  & 5{,}773  & 5{,}532 \\
    & Swap negatives        & 161{,}302 & 20{,}792 & 20{,}118 \\
    & Corruption negatives  & 64{,}679  & 8{,}369  & 8{,}051 \\
    & \textit{Total}        & 271{,}862 & 34{,}934 & 33{,}701 \\
    \midrule
    \multicolumn{5}{l}{\textit{Challenge Set (test-only)}} \\
    & Aligned / SAFE             & --- & --- & 685 \\
    & Aligned / SUSPICIOUS       & --- & --- & 252 \\
    & Aligned / MALICIOUS        & --- & --- & 213 \\
    & Misaligned / SAFE          & --- & --- & 196 \\
    & Misaligned / SUSPICIOUS    & --- & --- & 38 \\
    & Misaligned / MALICIOUS     & --- & --- & 60 \\
    & \textit{Total}             & --- & --- & 1{,}444 \\
    \bottomrule
    \end{tabular}
    \caption{Training and evaluation data counts. Corruption negatives aggregate the T3a/T3b/T3c variants. CPT and PL-HCL counts report the examples actually consumed during training and intrinsic evaluation after applying the 25\% per-split subsampling used for compute feasibility. Sequence-length budgets of 4,096 and 10,240 tokens were enforced at dataset-build time. The Challenge Set is used only for extrinsic evaluation. For the Challenge Set, alignment and maliciousness are treated as separate label axes; a skill can be aligned while still being suspicious or malicious if its surface claims accurately disclose the underlying behavior.}
    \label{tab:splits}
    \end{table*}

    \begin{table*}[!htbp]
    \centering
    \small
    \setlength{\tabcolsep}{6pt}
    \renewcommand{\arraystretch}{1.08}
    \begin{tabular}{lrrrr}
    \toprule
     & \textbf{Train} & \textbf{Val} & \textbf{Test} & \textbf{Unseen} \\
    \midrule
    \multicolumn{5}{l}{\textit{Whole corpus (skill packages)}} \\
    Collected + normalized corpus        & \multicolumn{4}{r}{264{,}937} \\
    After package-size filter ($\leq$256K chars) & \multicolumn{4}{r}{262{,}041} \\
    Challenge Set (human-reviewed, held out)     & \multicolumn{4}{r}{1{,}444} \\
    \midrule
    \multicolumn{5}{l}{\textit{Anchor packages --- Sub-stage 1 (metadata + instruction, $\leq$4{,}096 tok)}} \\
    Anchors in split \textit{(= CPT rows; T1/T2 cover all)} & 96{,}762 & 12{,}058 & 11{,}992 & 120{,}832 \\
    \quad with T3a behavior rewrite          & 96{,}371 & 12{,}021 & 11{,}948 & --- \\
    \quad with T3b donor span swap           & 96{,}316 & 12{,}012 & 11{,}935 & 120{,}303 \\
    \quad with T3c hallucinated identifiers  & 47{,}020 & 5{,}898  & 5{,}874  & --- \\
    \midrule
    \multicolumn{5}{l}{\textit{Anchor packages --- Sub-stage 2 (full package, $\leq$10{,}240 tok)}} \\
    Anchors in split \textit{(= CPT rows; T1/T2 cover all)} & 36{,}427 & 4{,}680 & 4{,}514 & 45{,}117 \\
    \quad with T3a behavior rewrite          & 36{,}421 & 4{,}678 & 4{,}513 & --- \\
    \quad with T3b donor span swap           & 36{,}417 & 4{,}677 & 4{,}512 & 45{,}105 \\
    \quad with T3c hallucinated identifiers  & 28{,}629 & 3{,}647 & 3{,}556 & --- \\
    \bottomrule
    \end{tabular}
    \caption{Whole-corpus and anchor-package accounting. Anchor counts denote distinct original skill packages that contribute at least one pair of the given type to the PL-HCL pool. Donors for swap/splice negatives are drawn within-pool, so the unseen pool is isolated from training. T3a excludes a small number of anchors whose rewrites were rejected or exceeded the token budget; T3c covers only anchors with substitutable real identifiers. Sub-stage 1 and Sub-stage 2 are materialized separately and partially overlap, rather than being cumulative.}
    \label{tab:anchor-accounting}
    \end{table*}

    \begin{table}[H]
    \centering
    \small
    \setlength{\tabcolsep}{5pt}
    \renewcommand{\arraystretch}{1.05}
    \begin{tabular}{lll}
    \toprule
    \textbf{Hyperparameter} & \textbf{Full-CPT} & \textbf{PL-HCL} \\
    \midrule
    Context length             & 4{,}096 / 10{,}240 & 4{,}096 / 10{,}240 \\
    Optimizer                  & AdamW              & AdamW \\
    Peak learning rate         & $2{\times}10^{-5}$ & $1{\times}10^{-5}$ \\
    Schedule                   & cosine, 3\% warmup & cosine, 3\% warmup \\
    Effective batch size       & 64 sequences       & 32 anchors $\times$ 5 views \\
    Weight decay               & 0.1                & 0.1 \\
    Precision                  & bf16               & bf16 \\
    Gradient clipping          & 1.0                & 1.0 \\
    Temperature $\tau$         & ---                & 0.07 \\
    Layer weights $\alpha_\ell$ & ---               & $(1/3,1/3,1/3)$ \\
    Pair-kind weights          & ---                & 1.0 : 1.5 : 1.0 \\
    Type-B corruption rate     & ---                & 0.5 / 0.3 / 0.2 \\
    Epochs                     & 1                  & 1 \\
    \bottomrule
    \end{tabular}
    \caption{Training hyperparameters for Full-CPT and PL-HCL. Context lengths and corruption rates correspond to the two-stage training schedule.}
    \label{tab:hyperparams}
    \end{table}

\section{Intrinsic Evaluation}
\label{app:intrinsic}

    We run two intrinsic diagnostics to verify that the pretrained backbones learn the intermediate objectives used in PL-HCL. First, next-token prediction (NTP) on held-out skill text assesses whether CPT improves the modeling of Agent Skill packages. Second, the held-out PL-HCL contrastive evaluation assesses whether the model separates aligned pairs from constructed swap and corruption negatives. These diagnostics support the main results in \S\ref{sec:results}: CPT improves skill-text modeling, and PL-HCL learns the contrastive objective, while extrinsic misalignment detection remains the primary evaluation.
    
    \paragraph{\textbf{1. Next-Token Prediction: }} Table~\ref{tab:intrinsic} reports NTP metrics before and after CPT. For both full-pipeline backbones, CPT Sub-stage 2 enhances final held-out skill-text metrics, reducing perplexity and boosting top-1 accuracy.
    
    \begin{table*}[!htbp]
    \centering
    \small
    \setlength{\tabcolsep}{5pt}
    \renewcommand{\arraystretch}{1.1}
    \begin{tabular}{llrrrrrr}
    \toprule
    \textbf{Backbone} & \textbf{Stage} & \textbf{Loss $\downarrow$} & \textbf{ppl$_{\text{word}}$ $\downarrow$} & \textbf{Top-1 $\uparrow$} & \textbf{Top-5 $\uparrow$} & \textbf{Top-$p$=0.95 $\uparrow$} & \textbf{BPB $\downarrow$} \\
    \midrule
    Foundation-Sec-8B-R   & base            & 1.188 & 10.679 & 0.725 & 0.886 & 0.959 & 0.426 \\
                          & + CPT sub-stg 1 & 1.402 & 13.127 & 0.678 & 0.861 & 0.971 & 0.485 \\
                          & + CPT sub-stg 2 & \textbf{1.164} & \textbf{8.904}  & \textbf{0.727} & \textbf{0.890} & \textbf{0.973} & \textbf{0.390} \\
    \midrule
    Llama-3.1-8B          & base            & 1.240 & 11.849 & 0.718 & 0.877 & 0.968 & 0.444 \\
                          & + CPT sub-stg 1 & 1.420 & 13.560 & 0.677 & 0.859 & 0.973 & 0.491 \\
                          & + CPT sub-stg 2 & \textbf{1.186} & \textbf{9.284}  & \textbf{0.724} & \textbf{0.887} & \textbf{0.975} & \textbf{0.397} \\
    \midrule
    Qwen-3-8B             & base            & 1.268 & 12.941 & 0.733 & 0.889 & 0.918 & 0.460 \\
    RedSage-Qwen3-8B-DPO  & base            & 1.160 & 10.398 & 0.731 & 0.887 & 0.956 & 0.421 \\
    WhiteRabbitNeo-2-8B   & base            & 1.181 & 10.524 & 0.726 & 0.884 & 0.971 & 0.423 \\
    \bottomrule
    \end{tabular}
    \caption{Intrinsic next-token-prediction metrics on held-out skill text. ppl$_{\text{word}}$ denotes word-level perplexity and BPB denotes bits-per-byte. Bold rows indicate the final post-CPT checkpoint.}
    \label{tab:intrinsic}
    \end{table*}

    \paragraph{\textbf{2. PL-HCL Contrastive Objective: }} Table~\ref{tab:hcl-combined} reports held-out PL-HCL performance by sub-stage and pair kind. The pair-kind decomposition shows that corrupted type-B pairs are more difficult than swapped type-A pairs, particularly for Foundation-Sec-8B-R, while Llama-3.1-8B performs strongly across all pair types.
    
\begin{table*}[t]
    \centering
    \small
    \setlength{\tabcolsep}{6pt}
    \renewcommand{\arraystretch}{1.05}
        \begin{subtable}[t]{0.50\textwidth}
        \centering
        \begin{tabular}{llcrrr}
        \toprule
        \textbf{Backbone} & \textbf{Stage} & \textbf{Split} & \textbf{Acc} & \textbf{F1} & \textbf{Gap} \\
        \midrule
        \multirow{4}{*}{FS-8B-R}
        & \multirow{2}{*}{sub-1} & test & 0.830 & 0.578 & 0.548 \\
        &                         & val  & 0.829 & 0.574 & 0.549 \\
        & \multirow{2}{*}{sub-2} & test & 0.867 & 0.708 & 0.629 \\
        &                         & val  & 0.863 & 0.703 & 0.623 \\
        \midrule
        \multirow{2}{*}{Qwen-3-8B}
        & \multirow{2}{*}{sub-1} & test & 0.994 & 0.976 & 0.993 \\
        &                         & val  & 0.995 & 0.980 & 0.994 \\
        \midrule
        \multirow{4}{*}{Llama-3.1-8B}
        & \multirow{2}{*}{sub-1} & test & 0.995 & 0.977 & 0.993 \\
        &                         & val  & 0.995 & 0.978 & 0.994 \\
        & \multirow{2}{*}{sub-2} & test & 0.986 & 0.959 & 0.979 \\
        &                         & val  & 0.984 & 0.955 & 0.977 \\
        \bottomrule
        \end{tabular}
        \caption{By PL-HCL sub-stage.}
        \label{tab:hcl-by-stage}
        \end{subtable}
    \hfill
        \begin{subtable}[t]{0.47\textwidth}
        \centering
        \begin{tabular}{llcrr}
        \toprule
        \textbf{Backbone} & \textbf{Stage} & \textbf{Pair kind} & \textbf{$n$} & \textbf{Acc} \\
        \midrule
        \multirow{6}{*}{FS-8B-R}
        & \multirow{3}{*}{sub-1} & positive  & 2{,}966  & 0.994 \\
        &                         & swapped   & 11{,}923 & 0.994 \\
        &                         & corrupted & 10{,}455 & 0.597 \\
        & \multirow{3}{*}{sub-2} & positive  & 5{,}332  & 0.984 \\
        &                         & swapped   & 20{,}118 & 0.940 \\
        &                         & corrupted & 8{,}051  & 0.602 \\
        \midrule
        \multirow{6}{*}{Llama-3.1-8B}
        & \multirow{3}{*}{sub-1} & positive  & 2{,}966  & 1.000 \\
        &                         & swapped   & 11{,}923 & 0.998 \\
        &                         & corrupted & 10{,}455 & 0.989 \\
        & \multirow{3}{*}{sub-2} & positive  & 5{,}332  & 0.996 \\
        &                         & swapped   & 20{,}118 & 0.992 \\
        &                         & corrupted & 8{,}051  & 0.964 \\
        \bottomrule
        \end{tabular}
        \caption{By pair kind.}
        \label{tab:hcl-by-kind}
        \end{subtable}

    \caption{Held-out PL-HCL objective by sub-stage and pair kind. Positive pairs are expected to score high, while swapped and corrupted pairs are expected to score low. \textit{Gap} denotes the positive mean probability minus the negative mean probability.}
    
    \label{tab:hcl-combined}
\end{table*}

\section{Supplementary Results of Full Evaluation}
    \label{app:metrics}
    
    Table~\ref{tab:supp-metrics} reports supplementary per-class metrics and confusion counts for the Challenge Set evaluation. These metrics complement the macro-F1 results reported in \S\ref{sec:results}. Because the Challenge Set is class-imbalanced, aligned-class performance alone can obscure failures on the misaligned class. We therefore report precision and recall for both classes, together with confusion counts using aligned as the positive class. Under this convention, \textit{false positives aligned} correspond to missed misaligned skills, while \textit{true negatives aligned} correspond to correctly detected misaligned skills.
    
    \begin{table*}[!htbp]
    \centering
    \small
    \setlength{\tabcolsep}{5pt}
    \renewcommand{\arraystretch}{1.05}
    \begin{tabular}{llcrrrrrr}
    \toprule
    \textbf{Model} & \textbf{Adapter} & \textbf{$k$}
    & \textbf{Prec$_a$} & \textbf{Rec$_a$}
    & \textbf{Prec$_m$} & \textbf{Rec$_m$}
    & \textbf{Missed$_m$} & \textbf{Detected$_m$} \\
    \midrule
    Llama-3.1-8B & raw & 0 & 0.797 & 0.989 & 0.278 & 0.017 & 289 & 5 \\
    Llama-3.1-8B & CPT-only & 0 & 0.807 & 0.993 & 0.750 & 0.071 & 273 & 21 \\
    Llama-3.1-8B & CPT+HCL & 0 & 0.856 & 1.000 & 1.000 & 0.344 & 193 & 101 \\
    Llama-3.1-8B & raw & 2 & 0.809 & 0.999 & 0.958 & 0.078 & 271 & 23 \\
    Llama-3.1-8B & CPT-only & 2 & 0.809 & 1.000 & 1.000 & 0.078 & 271 & 23 \\
    Llama-3.1-8B & CPT+HCL & 2 & 0.918 & 1.000 & 1.000 & 0.650 & 103 & 191 \\
    Llama-3.1-8B & raw & 5 & 0.797 & 1.000 & 1.000 & 0.003 & 293 & 1 \\
    Llama-3.1-8B & CPT-only & 5 & 0.805 & 1.000 & 1.000 & 0.051 & 279 & 15 \\
    Llama-3.1-8B & CPT+HCL & 5 & 0.915 & 1.000 & 1.000 & 0.636 & 107 & 187 \\
    \midrule
    Foundation-Sec-8B & raw & 0 & 0.797 & 0.989 & 0.235 & 0.014 & 290 & 4 \\
    Foundation-Sec-8B & CPT-only & 0 & 0.807 & 0.990 & 0.647 & 0.075 & 272 & 22 \\
    Foundation-Sec-8B & CPT+HCL & 0 & 0.877 & 1.000 & 1.000 & 0.452 & 161 & 133 \\
    Foundation-Sec-8B & raw & 2 & 0.809 & 0.997 & 0.852 & 0.078 & 271 & 23 \\
    Foundation-Sec-8B & CPT-only & 2 & 0.810 & 0.998 & 0.923 & 0.082 & 270 & 24 \\
    Foundation-Sec-8B & CPT+HCL & 2 & 0.927 & 1.000 & 1.000 & 0.690 & 91 & 203 \\
    Foundation-Sec-8B & raw & 5 & 0.797 & 0.995 & 0.250 & 0.007 & 292 & 2 \\
    Foundation-Sec-8B & CPT-only & 5 & 0.803 & 0.997 & 0.750 & 0.041 & 282 & 12 \\
    Foundation-Sec-8B & CPT+HCL & 5 & 0.925 & 1.000 & 1.000 & 0.684 & 93 & 201 \\
    \midrule
    WhiteRabbitNeo-2-8B & raw & 0 & 0.797 & 0.990 & 0.267 & 0.014 & 290 & 4 \\
    WhiteRabbitNeo-2-8B & raw & 2 & 0.805 & 0.994 & 0.708 & 0.058 & 277 & 17 \\
    WhiteRabbitNeo-2-8B & raw & 5 & 0.805 & 0.997 & 0.857 & 0.061 & 276 & 18 \\
    \midrule
    RedSage-Qwen3-8B-DPO & raw & 0 & 0.801 & 0.997 & 0.750 & 0.031 & 285 & 9 \\
    RedSage-Qwen3-8B-DPO & raw & 2 & 0.808 & 0.997 & 0.875 & 0.071 & 273 & 21 \\
    RedSage-Qwen3-8B-DPO & raw & 5 & 0.809 & 0.998 & 0.920 & 0.078 & 271 & 23 \\
    \midrule
    Qwen-3-8B & raw & 0 & 0.800 & 0.997 & 0.727 & 0.027 & 286 & 8 \\
    Qwen-3-8B & raw & 2 & 0.809 & 0.997 & 0.889 & 0.082 & 270 & 24 \\
    Qwen-3-8B & raw & 5 & 0.809 & 0.996 & 0.821 & 0.078 & 271 & 23 \\
    \bottomrule
    \end{tabular}
    \caption{Supplementary per-class metrics and misalignment counts on the Challenge Set. Prec$_a$ and Rec$_a$ denote aligned-class precision and recall; Prec$_m$ and Rec$_m$ denote misaligned-class precision and recall. Missed$_m$ denotes misaligned skills incorrectly predicted as aligned; Detected$_m$ denotes misaligned skills correctly predicted as misaligned.}
    \label{tab:supp-metrics}
    \end{table*}

\end{document}